\documentclass{article}

\usepackage[preprint, nonatbib]{neurips_2026}
\usepackage[
    backend=biber,
    style=apa,
    natbib=true
]{biblatex}

\usepackage[utf8]{inputenc} 
\usepackage[T1]{fontenc}    
\usepackage{hyperref}       
\usepackage{url}            
\usepackage{booktabs}       
\usepackage{amsfonts}       
\usepackage{nicefrac}       
\usepackage{microtype}      
\usepackage{xcolor}         
\usepackage{graphicx}
\usepackage{multirow}
\usepackage{float}
\usepackage[normalem]{ulem}
\usepackage{amsmath} 
\usepackage{subcaption}
\usepackage{tabularx}
\usepackage{array}
\usepackage{makecell}
\usepackage{etoolbox}
\usepackage{setspace}

\author{%
Chaeho Lee$^{1}$,
Junyup Kim$^{1}$,
Wansoo Kim$^{2}$,
Young Min Jung$^{1}$,
Sang Ho Lee$^{1,2,3\, \dagger}$
\\
$^{1}$School of Digital Humanities and Computational Social Sciences, KAIST
\\
$^{2}$Department of Brain \& Cognitive Sciences, KAIST
\\
$^{3}$Graduate School of Data Science, KAIST
\\
\texttt{\{chaeholee, ytrewq271828, wsk37, youngmin, sangholee\}@kaist.ac.kr}
}
\begin{document}

\title{Leveraging Psychophysical Attentional Distribution for Gaze-Augmented Reward Modeling}
\maketitle
\begingroup
\renewcommand{\thefootnote}{\ensuremath{\dagger}}
\footnotetext{Corresponding author.}
\endgroup

\begin{abstract}
    Integrating human feedback into language models has gained increasing attention as a means to align model outputs with human preferences, with Reinforcement Learning from Human Feedback (RLHF) serving as a prominent framework for preference-based alignment. Recent studies have incorporated eye-tracking (ET) data as an additional supervisory signal for reward model training, grounding reward learning in human reading behavior. Building on this approach, we extend prior fixation-centric models by leveraging established findings from cognitive psychology on human letter recognition. Specifically, we model visual attention as a spatially graded distribution centered on the current gaze location and incorporate this representation into reward model training. Our results show that the reward model trained with the proposed gaze distribution achieves higher preference prediction accuracy than a baseline model. Moreover, some results from our experiments are qualitatively consistent with established findings on human attention during reading, suggesting that they potentially capture cognitively meaningful aspects of human attention allocation.
    
\end{abstract}

\section{Introduction}
Reinforcement Learning from Human Feedback (RLHF) is a widely used paradigm for aligning large language models (LLMs) with human preferences. In a typical RLHF pipeline, a reward model (RM) is trained to predict human preferences over pairs of responses from a generative LLM by assigning a higher reward to a chosen response than to a rejected one. Since the RM serves as a proxy for human judgment, its ability to faithfully capture human preferences is central to alignment quality.

A limitation of standard RLHF, however, is that such supervision is typically restricted to the final preference judgment. The model observes which response a human prefers, but not how the human cognitively processes the responses when deciding their preferred response. This limitation points to an opportunity to enrich reward modeling with human cognitive signals that arise during text processing. Among such signals, gaze signals derived from eye-tracking data are particularly relevant because gaze behavior reflects how readers allocate visual attention and cognitive effort across a text \citep{just1980theory}. These data can be represented as gaze features (e.g., gaze location, fixation duration) mapped to individual words or tokens, and such gaze features can be used as additional inputs for reward modeling.

Recent studies suggest that gaze features can improve reward modeling for language model alignment. For example, \citet{lopez2025seeing} found that GazeReward, a gaze-augmented reward modeling framework that incorporates gaze features as additional inputs, increased the accuracy of human preference prediction.  \citet{galliamov2025enhancing} further investigated gaze-informed reward modeling and showed that incorporating gaze features can improve alignment efficiency while maintaining performance.

Although these studies highlight the potential of gaze features for reward modeling, their approaches underrepresent a key property of human information processing during reading. They represent human gaze as a discrete token-wise feature associated with the directly fixated token. However, humans can acquire visual information beyond the point of fixation, because visual attention is spatially distributed rather than confined to a fixation  \citep{schotter2012parafoveal}. The spatial region from which such information can be acquired during a fixation is often referred to as the perceptual span. In left-to-right reading, this region typically extends further to the right of fixation than to the left \citep{rayner2014gaze}.

Based on these findings, we hypothesize that redistributing token-wise gaze features across neighboring tokens according to an asymmetric profile will improve reward modeling by capturing aspects of human cognitive processing relevant to preference judgment. Specifically, we apply a profile inspired by the spatial distribution of human attention, assigning lower weights to tokens farther from the fixated token. The spread of the distribution on each side of fixation is controlled by separate parameters learned during reward model training. Across experimental settings, our approach improves preference prediction accuracy relative to a baseline. We also find that the optimized shape of the asymmetric distribution tends to extend more broadly to the right than to the left, which is consistent with the perceptual span reported in psychophysical research.

\section{Background and Related Work}

\subsection{Visual Information Processing and Attention During Reading}
Human reading is not a process in which visual information is acquired from one isolated word at a time. Instead, reading proceeds through a sequence of fixations, and readers obtain information from a spatially limited region around each fixation. Importantly, this region is not confined to the directly fixated word. Although visual processing is most precise at the fixated word, readers also extract useful information from neighboring text \citep{schotter2012parafoveal}. This suggests that a computational representation of gaze should extend across the region surrounding the fixated token to better reflect human information processing.

\begin{figure}
    \centering
    \includegraphics[width=\linewidth]{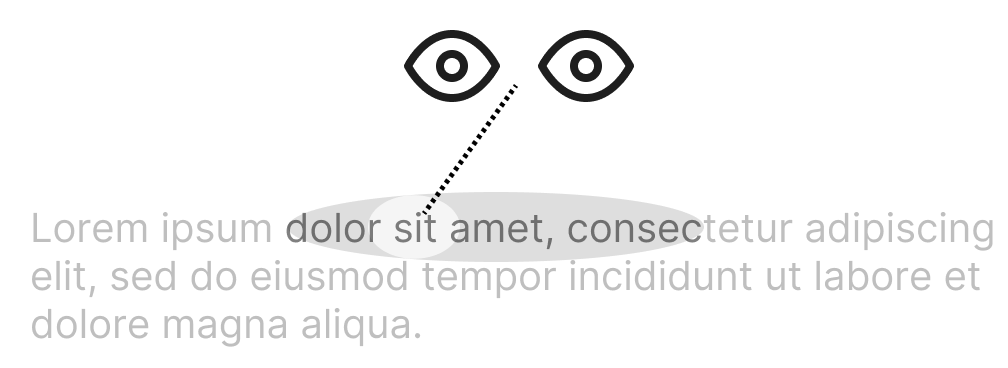}
    \caption{A schematic illustration of visual attention allocation during each fixation while reading.}
    \label{fig:schematic}
\end{figure}

A study on the visual span during reading provides a more detailed account of how visual processing varies across the region surrounding fixation. \citet{legge2001psychophysics} found that letter-recognition accuracy within the visual span is spatially graded, with accuracy being highest near fixation and decreasing with distance from fixation. They modeled this pattern as an asymmetric Gaussian function, allowing the spread of recognition accuracy to differ between the left and right sides of fixation. This asymmetric function is analogous to a spatial distribution of visual attention, assigning different weights to locations around fixation. In left-to-right reading, visual attention is consistently found to be asymmetric, extending only about 3–4 letter spaces to the left of fixation but about 14–15 letter spaces to the right \citep{rayner2014gaze}. Figure~\ref{fig:schematic}  schematically illustrates this asymmetric attentional distribution during reading.

Computational models of reading also describe how human visual attention guides word recognition and eye movements across text. For example, sequential  attention shift (SAS; \citealp{inhoff2005time}) models, such as E-Z Reader \citep{reichle1998toward, reichle2006ez, reichle2011serial}, describe attention as shifting from one word to the next as lexical processing progresses. In contrast, processing gradient (PG; \citealp{inhoff2005time}) models, such as SWIFT \citep{engbert2005swift}, Glenmore \citep{reilly2006some}, and OB1-reader \citep{snell2018ob1}, assume parallel processing of multiple adjacent words under a spatially distributed attentional gradient. Notably, OB1-reader is particularly relevant to the current approach because it assumes an asymmetric attentional distribution.

\subsection{Gaze-Augmented Natural Language Processing}
Gaze information from eye-tracking data has been increasingly used to improve various downstream natural language processing (NLP) tasks. \citet{mishra2016leveraging} augmented traditional linguistic features with gaze-derived cognitive features for sentiment analysis and sarcasm detection. \citet{sood2020improving} and \citet{zhang2024eyetrans} incorporated gaze information into neural attention mechanisms for paraphrase generation and code summarization, respectively. Together, these studies show that gaze features can be a useful resource for improving model performance across NLP tasks.

Recent research has extended the use of gaze information to language model post-training. For instance, \citet{kiegeland2024pupil} incorporated human eye-tracking data as direct feedback within Direct Preference Optimization (DPO). In addition, \citet{deng2024fine} proposed a fine-tuning framework that reorders token embeddings based on simulated gaze trajectories. Gaze prediction models such as ScanTextGAN \citep{khurana2023synthesizing} simulate human-like gaze patterns for new texts, enabling the use of gaze features without collecting additional eye-tracking data. GazeReward \citep{lopez2025seeing}, which serves as the foundation of our approach, uses gaze prediction models proposed by \citet{huang2023long} and \citet{li2021torontocl}.

\subsection{GazeReward}
In this section, we provide an overview of GazeReward \citep{lopez2025seeing}, the gaze-augmented reward modeling framework on which our approach directly builds. 

GazeReward extends standard reward modeling by incorporating eye-tracking (ET) features as additional inputs to the reward model. In conventional reward modeling, the model receives a prompt-response pair and learns to assign a higher scalar reward to the response preferred by human annotators. GazeReward keeps this basic preference-learning objective, but augments the input representation with gaze-derived features. 

Because human ET data are costly to collect and unavailable for most preference datasets, GazeReward uses pretrained ET prediction models to generate synthetic gaze features for each prompt-response pair. Given the same text that is passed to the reward model, the ET generator predicts token-level gaze features such as total reading time (TRT), first fixation duration (FFD), go-past time (GPT), number of fixations (nFix), and fixation proportion (fixProp). These features are then projected into the reward model’s embedding space through a neural feature projector, allowing gaze-derived information to be combined with the model’s text embeddings.

GazeReward considers two strategies for incorporating ET features into the reward model. In the GazeConcat setting, the projected ET embeddings are concatenated with the text embeddings, with special separator tokens marking the gaze feature segment. In the GazeAdd setting, the projected ET embeddings are added directly to the corresponding text embeddings. Since the tokenizer used by the ET prediction model may differ from the tokenizer used by the reward model, GazeReward also performs token remapping so that gaze features can be aligned with the reward model’s input tokens before they are combined with text representations.

After constructing the gaze-augmented input, the reward model is trained with the standard pairwise preference loss. For a prompt and two candidate responses, the model produces scalar rewards for the chosen and rejected responses, and training encourages the reward for the chosen response to be higher than that for the rejected response.

\section{Methods}
\begin{figure}
    \centering
    \includegraphics[width=\linewidth]{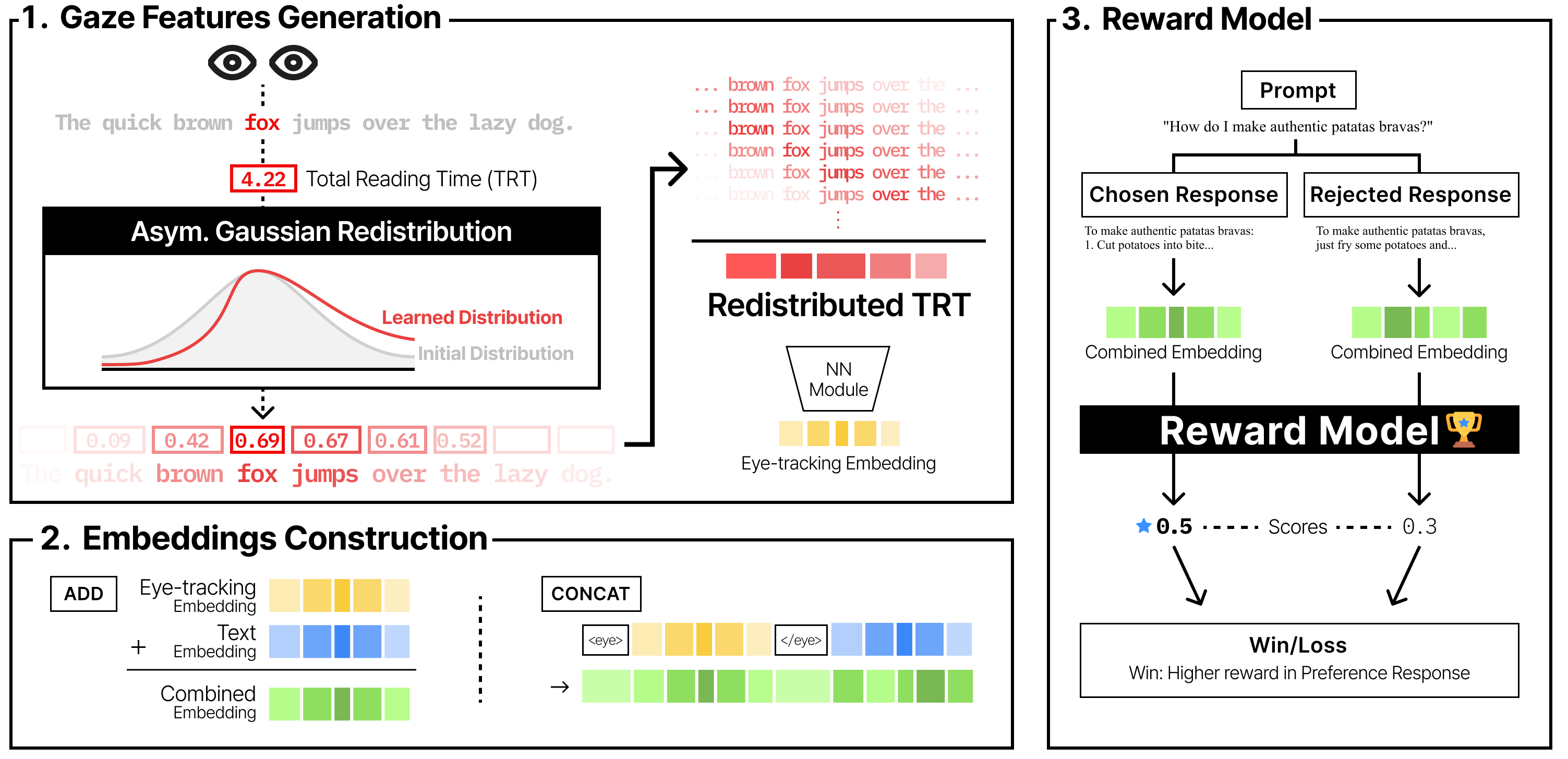}
    \caption{An overview of the GazeReward framework with our redistribution module. First, we obtain original TRT values for a given sentence and redistribute them using an asymmetric Gaussian kernel with learnable parameters. The TRT values are then transformed into eye-tracking embeddings through the ET feature projector.}
    \label{fig:pipeline}
\end{figure}

Our method modifies the gaze-feature construction stage of GazeReward. In the original GazeReward
pipeline, token-level ET features are first generated by a pretrained ET generator and then directly
projected into the reward model’s embedding space through the ET feature projector. In contrast, we
insert an additional redistribution module between these two steps. Figure~\ref{fig:pipeline} shows the implementation of our redistribution module in the GazeReward pipeline.

GazeReward addresses five eye tracking features. Among these features, we use total reading time (TRT) as the core input to our formulation. TRT measures the cumulative duration of all fixations on a word and thus reflects the total attention allocated by readers.

Specifically, after obtaining token-level TRT values from the pretrained ET generator model proposed by \citet{huang2023long}, we redistribute each token’s TRT value across its neighboring tokens using an asymmetric Gaussian kernel. A sequence of redistributed TRT values is then passed to the ET feature projector, which generates the eye-tracking embedding. The eye-tracking embedding is subsequently concatenated with (GazeConcat) or added to (GazeAdd) text embeddings. Our method does not alter the pairwise reward modeling objective or the overall reward model architecture; it changes the representation of gaze features before they are incorporated into the reward model.

\begin{figure}[h]
    \centering
    \includegraphics[width=0.75\linewidth]{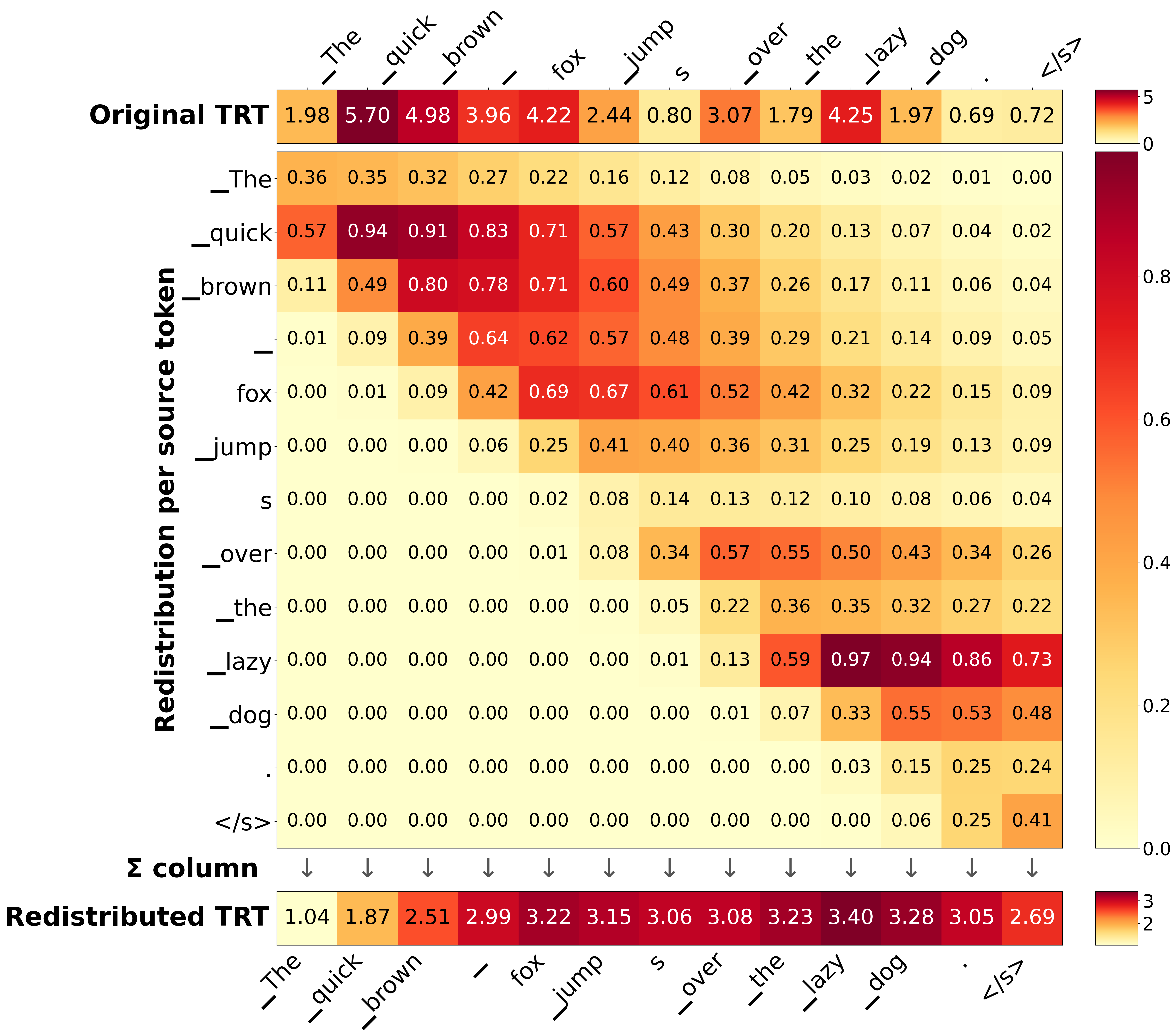}
    \caption{Asymmetric Gaussian redistribution of total reading time (TRT). Example of redistributing per-token TRT values for the sentence ``The quick brown fox jumps over the lazy dog.'' (Top) Original TRT values predicted per token. (Middle) Each cell indicates how much of source token's TRT is redistributed to neighboring tokens weighted by an asymmetric Gaussian kernel with $\sigma_\text{left} = 1.0$ and $\sigma_\text{right} = 4.0$. (Bottom) Redistributed TRT values obtained by summing each column of the contribution matrix.}
    \label{fig:asym_redist}
\end{figure}

While TRT provides information about which tokens were directly fixated, human text recognition is not confined to the fixated region alone. Cognitive and psychophysical studies of reading suggest that visual information is acquired over a visual span encompassing both central and peripheral vision
\citep{legge2001psychophysics}. Accordingly, we redistribute token-level 
TRT values to adjacent tokens to reflect this perceptual span.

Formally, let $\{\mathrm{tok}_1, \dots, \mathrm{tok}_S\}$ denote a sequence of $S$ tokens, and 
let $\mathrm{TRT}_j$ be the predicted TRT associated with token $\mathrm{tok}_j$. For each 
fixation at $\mathrm{tok}_j$, we define a perceptual field over neighboring tokens using 
an asymmetric Gaussian kernel centered at $\mathrm{tok}_j$, where distances are measured 
in token indices. The weight from fixation $j$ to token $\mathrm{tok}_i$ is:
\begin{equation}
    k_{j \rightarrow i} = \exp\!\left(-\frac{(i-j)^2}{2\,{\sigma^{(\mathrm{side})}}^2}\right),
\end{equation}
where $\sigma^{(\mathrm{side})}$ is $\sigma_{\mathrm{left}}$ when $i < j$ and 
$\sigma_{\mathrm{right}}$ when $i \geq j$, reflecting the asymmetric nature of 
human reading. Both $\sigma_{\mathrm{left}}$ and $\sigma_{\mathrm{right}}$ are 
learnable parameters jointly optimized with the reward model during training, 
allowing the perceptual span to adapt to the downstream alignment objective 
rather than being fixed a priori.

The weights are normalized per fixation to preserve the total TRT mass:
\begin{equation}
    \hat{k}_{j \rightarrow i} = \frac{k_{j \rightarrow i}}{\displaystyle\sum_{i'=1}^{S} k_{j \rightarrow i'}}.
\end{equation}
The final redistributed TRT for token $\mathrm{tok}_i$ aggregates contributions from all fixated tokens:
\begin{equation}
    g_i = \sum_{j=1}^{S} TRT_j \cdot \hat{k}_{j \rightarrow i}.
\end{equation}
The resulting sequence $\{\mathrm{g}_1, \dots, \mathrm{g}_S\}$ forms a dense token-level TRT distribution that captures both direct fixations and peripheral information acquired during reading.

Figure~\ref{fig:asym_redist} illustrates this redistribution process using a simple sentence as an example. The top row shows the original token-level TRT values predicted by the ET generator. The middle matrix shows how each source token distributes its TRT value to neighboring target tokens according to the asymmetric Gaussian kernel. Each row therefore represents the redistribution pattern from a single source token. The bottom row shows the final redistributed TRT values obtained by summing the contributions in each column. This example visualizes how the proposed module transforms sparse token-local TRT values into a smoother, spatially distributed gaze representation before projection into the reward model.

\section{Experiments}

Our experiments are designed to address two central questions. First, we examine whether redistributing token-level TRT according to an asymmetric distribution assumption improves reward modeling accuracy over the baseline, which uses token-wise gaze representation. Second, we analyze whether the optimized asymmetric Gaussian parameters $\sigma_\text{left}$ and $\sigma_\text{right}$ show a directional pattern consistent with human visual attention during left-to-right reading. 

\subsection{Experimental Setup}
\subsubsection{Practical Adjustments and Datasets}
We run our experiments on the OpenAssistant Conversations dataset (OASST1; \citealp{kopf2023openassistant}), HelpSteer2 \citep{wang2024helpsteer}, and UltraFeedback \citep{cui2024ultrafeedback}. Due to consistency with the publicly released code of the original GazeReward implementation and to our computational resource constraints, we cap the maximum input length across the datasets and exclude samples exceeding this threshold. Table~\ref{tab:dataset_length_cap} shows the input length cap imposed on each dataset and the corresponding amount of data loss.

\begin{table}[H]
    \centering
    \caption{Data loss under the input length cap across the three datasets.}
    \label{tab:dataset_length_cap}
    \begin{tabular}{lccc}
        \toprule
        & OASST1 & HelpSteer2 & UltraFeedback \\
        \midrule
        Input Length Cap (characters) & 5,000 & 10,000 & 10,000 \\
        Total Data Samples & 8,161 & 7,592  & 63,966 \\
        Data Loss & 307 (3.76\%) & 241 (3.17\%) & 1,632 (2.55\%) \\
        \bottomrule
    \end{tabular}
\end{table}

Additionally, we set the per-device batch size to 1 and use gradient accumulation over 8 steps, preserving an effective batch size of 8, which matches the batch size used in the prior study.

\subsubsection{Model Choice and Gaze Integration Settings}
For the experiments on OASST1, we focus on two variants of Llama-3-based reward models: Llama-3-8B and Llama-3-8B-Instruct \citep{grattafiori2024llama}. We include both variants to assess whether the effect of the redistribution is consistent across the two Llama-3 initializations, rather than restricting the evaluation to a single model variant. We also evaluate both gaze integration settings (GazeConcat and GazeAdd) introduced in GazeReward.

\subsubsection{Training Configurations for Redistribution Parameters}
The $\sigma_\text{left}$ and $\sigma_\text{right}$ parameters are initialized to 1.0 and optimized jointly with the base parameters during training. Because these parameters operate at a different scale from the main reward model parameters, we use a separate learning rate of $5 \times 10^{-2}$ for $\sigma_\text{left}$ and $\sigma_\text{right}$, while the remaining model parameters are trained with a learning rate of $5 \times 10^{-5}$. These hyperparameter values were selected through empirical tuning. After training, we analyze the $\sigma_\text{left}$ and $\sigma_\text{right}$ values from the best checkpoint.

For each experimental condition, we conduct six runs using different random seeds and report the mean reward modeling accuracy with the standard deviation. Further details on the experimental setups are provided in Appendix~\ref{app:experiment_details}.

\section{Results}

Table~\ref{tab:oasst1_results} reports the reward modeling accuracy on OASST1 across six random seeds. 
Overall, our approach improves performance over the corresponding baseline in all four experimental settings. 
In the GazeConcat setting, our approach achieves 74.73\% accuracy with Llama-3-8B-Instruct, compared to 68.38\% for the baseline, corresponding to a 9.29\% relative improvement. For Llama-3-8B, the GazeConcat setting also improves from 69.23\% to 74.81\%, yielding an 8.06\% relative improvement. These results indicate that asymmetrically redistributing gaze features can improve reward preference prediction across both gaze integration methods and both Llama-3 model variants.

\begin{table}[H]
\centering
\caption{Mean reward modeling accuracy on OASST1 across six random seeds. Diff. denotes the relative improvement of our approach over the corresponding baseline (mean $\pm$ standard deviation).}
\label{tab:oasst1_results}
\begin{tabular}{llcccc}
\toprule
& & \multicolumn{2}{c}{\textbf{Llama-3-8B-Instruct}}
& \multicolumn{2}{c}{\textbf{Llama-3-8B}} \\
\cmidrule(lr){3-4} \cmidrule(lr){5-6}
& & Accuracy (\%) & Diff. (\%) & Accuracy (\%) & Diff. (\%) \\
\midrule
\multirow{2}{*}{\textbf{GazeConcat}}
& Baseline & 68.38 $\pm$ 3.88 & -- & 69.23 $\pm$ 4.92 & -- \\
& Our Approach & 74.73 $\pm$ 1.55 & 9.29 & 74.81 $\pm$ 1.32 & 8.06 \\
\midrule
\multirow{2}{*}{\textbf{GazeAdd}}
& Baseline & 70.30 $\pm$ 1.25 & -- & 69.27 $\pm$ 0.81 & -- \\
& Our Approach & 71.89 $\pm$ 1.22 & 2.26 & 71.17 $\pm$ 1.21 & 2.74 \\
\bottomrule
\end{tabular}
\end{table}

Table~\ref{tab:sigma_values} summarizes the asymmetric Gaussian parameters observed at the best checkpoints. 
In most settings, the right-side spread is larger than the left-side spread, indicating that the optimized redistribution tends to assign a wider influence range to tokens on the right side of the fixated point. 
This tendency is especially clear in the GazeConcat setting, where $\sigma_{\mathrm{right}}$ exceeds $\sigma_{\mathrm{left}}$ for both Llama-3-8B-Instruct and Llama-3-8B. The GazeAdd setting shows a similar pattern for Llama-3-8B-Instruct, while the difference is smaller for Llama-3-8B.

\begin{table}[H]
\centering
\caption{Asymmetric Gaussian parameters from the best checkpoints on OASST1 across six random seeds (mean $\pm$ standard deviation).}
\label{tab:sigma_values}
\begin{tabular}{lcccc}
\toprule
& \multicolumn{2}{c}{\textbf{Llama-3-8B-Instruct}}
& \multicolumn{2}{c}{\textbf{Llama-3-8B}} \\
\cmidrule(lr){2-3} \cmidrule(lr){4-5}
& $\sigma_{\mathrm{left}}$ & $\sigma_{\mathrm{right}}$
& $\sigma_{\mathrm{left}}$ & $\sigma_{\mathrm{right}}$ \\
\midrule
\textbf{GazeConcat}
& 1.62 $\pm$ 1.18 & 2.25 $\pm$ 1.71
& 0.80 $\pm$ 0.56 & 3.47 $\pm$ 2.13 \\
\midrule
\textbf{GazeAdd}
& 0.58 $\pm$ 0.21 & 1.99 $\pm$ 1.17
& 0.64 $\pm$ 0.23 & 0.80 $\pm$ 0.31 \\
\bottomrule
\end{tabular}
\end{table}

\begin{figure}[!b]
    \centering
    \includegraphics[width=1\linewidth]{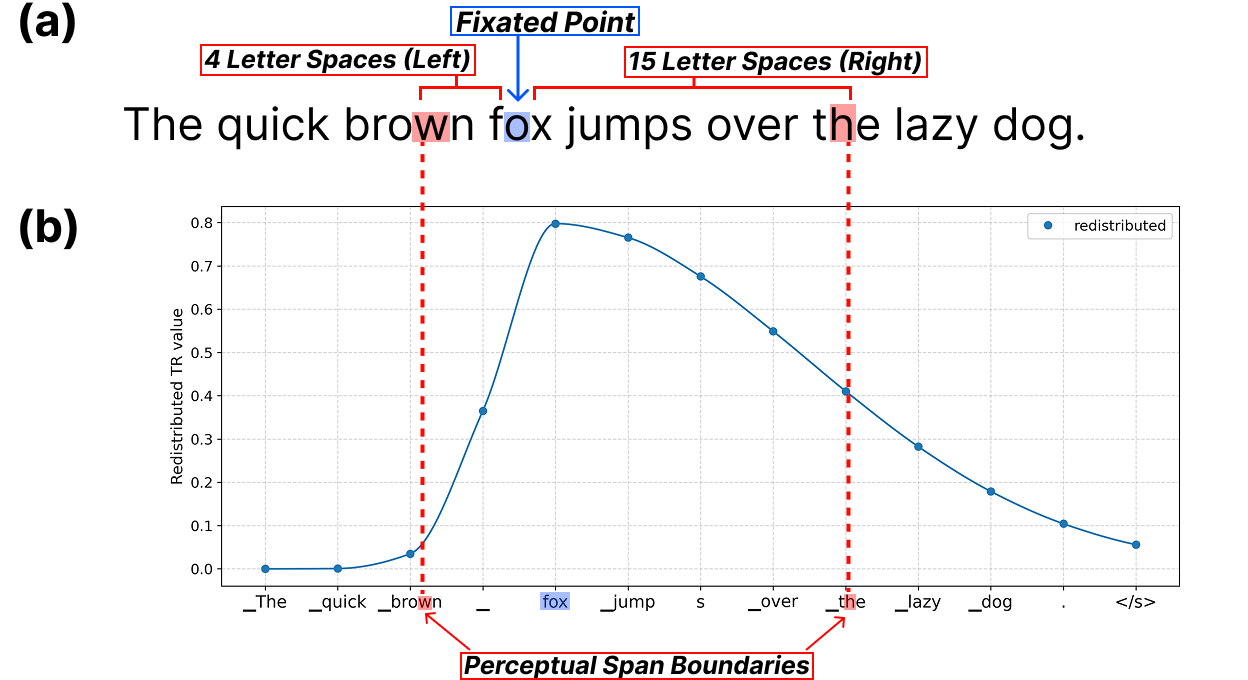}
    \caption{An illustration of the human perceptual span and the proposed token-level gaze redistribution. (a) The asymmetric human perceptual span in left-to-right reading. (b) TRT values redistributed to the tokens adjacent to the token \textit{fox}. The red dotted lines indicate the hypothetical boundaries of the human perceptual span mapped onto the tokenized sentence.}
    \label{fig:rayner_span}
\end{figure}

Figure~\ref{fig:rayner_span} visualizes the token-level distribution of attention defined by the estimated $\sigma_{\text{left}}$ and $\sigma_{\text{right}}$, over a schematic perceptual span during reading. The upper panel (a) shows a perceptual span around the fixated point (on the character \textit{o} in the word \textit{fox}) based on the finding that the span extends 4 letter spaces to the left (up to the character \textit{w} in the word \textit{brown}) and 15 letter spaces to the right (up to the character \textit{h} in the word \textit{the}) from the fixated point during left-to-right reading \citep{rayner2014gaze}. The lower panel (b) illustrates the asymmetric Gaussian distribution derived from the best-performing Llama-3-8B model in the GazeConcat setting, where $\text{mean }\sigma_{\text{left}}=0.80$ and $\text{mean }\sigma_{\text{right}}=3.47$. The blue curved line shows the redistributed TRT values across the adjacent tokens given that the token \textit{fox} has a TRT value of 4.221. This example is not intended as a direct quantitative mapping from letter spaces to tokens, but as a qualitative illustration of the learned redistribution pattern across letters.

We further evaluate our method on two additional datasets, HelpSteer2 and UltraFeedback, using Llama-3-8B as the reward model backbone and GazeConcat as the gaze integration method, as this configuration achieved the highest accuracy on OASST1. Table ~\ref{tab:cross_dataset_results} summarizes the results.

\begin{table}[H]
\centering
\caption{Reward modeling performance on additional preference datasets.
Results are reported as mean $\pm$ standard deviation across
six random seeds. Diff. denotes the relative improvement of our approach
over the corresponding baseline.}
\label{tab:cross_dataset_results}
\begin{tabular}{llcccc}
\toprule
\textbf{Dataset} & \textbf{Method} & \textbf{Accuracy (\%)}
& \textbf{Diff. (\%)}
& $\boldsymbol{\sigma}_{\mathrm{left}}$
& $\boldsymbol{\sigma}_{\mathrm{right}}$ \\
\midrule
\multirow{2}{*}{HelpSteer2}
& Baseline
& $59.48 \pm 3.66$
& --
& --
& -- \\
& Our Approach  
& $62.64 \pm 1.68$
& $5.31$ 
& $3.18 \pm 2.34$
& $1.85 \pm 2.01$ \\
\midrule
\multirow{2}{*}{UltraFeedback}
& Baseline
& $65.65 \pm 2.12$
& --
& --
& -- \\
& Our Approach 
& $74.20 \pm 0.46$ 
& $13.02$
& $1.48 \pm 1.69$ 
& $3.79 \pm 1.94$ \\
\bottomrule
\end{tabular}
\end{table}

Our approach improves reward model performance on both additional datasets. On HelpSteer2, our approach achieves 62.64\% accuracy compared with 59.48\% for the baseline, corresponding to a 5.31\% relative improvement. A substantially larger gain is observed on UltraFeedback, where accuracy increases from 65.65\% to 74.20\%, yielding a 13.02\% relative improvement.

HelpSteer2 showed the smallest improvement among the datasets and was the only dataset on which the learned profile extended further to the left than to the right. We speculate that the differences in the number of candidate responses available for chosen--rejected pair construction may help explain the distinct results on HelpSteer2. In our processed data, OASST1 contains an average of 2.82 candidate responses per prompt, while UltraFeedback contains 4 responses for nearly every prompt. For these datasets, our preprocessing selects the highest- and lowest-rated responses as the chosen--rejected pair. By contrast, HelpSteer2 provides only 2 responses per prompt in almost all cases, and our preprocessing designates the response with the higher rating as chosen and the other as rejected. Thus, HelpSteer2 generally requires using the only available pair, whereas OASST1 and UltraFeedback allow selecting maximally separated pairs from larger candidate pools. This constraint may yield weaker preference contrasts, potentially making preference prediction more challenging and providing a less informative training signal for learning the redistribution profile.

\section{Ablation Study}
In this section, we examine the contributions of different design choices in our method, using Llama-3-8B as the reward model backbone in the GazeConcat setting on the OASST1 dataset. We compare the original pinpoint representation against uniform redistribution, a fixed asymmetric Gaussian, a learnable symmetric Gaussian, and our learnable asymmetric Gaussian across six random seeds. Table ~\ref{tab:ablation} presents the results.

Our ablation study supports the following observations. First, the fixed 1:4 condition achieved higher mean accuracy (71.46\%) than the uniform 1:4 condition (70.18\%). Although the difference was not large enough to draw a firm conclusion given the variability across seeds, the results suggest that implementing a Gaussian-shaped distribution may be more effective than uniformly distributing the TRT values over the neighboring tokens.

Second, the learnable symmetric and asymmetric Gaussian conditions achieve higher accuracy than the uniform and fixed Gaussian conditions. These results support the importance of learning the redistribution parameters for achieving the improvement.

\begin{table}[H]
    \centering
    \caption{Reward modeling accuracy across five conditions: Pinpoint baseline, Uniform 1:4, Fixed 1:4, Symmetric, and Asymmetric. A description for each condition and its accuracy (mean $\pm$ standard deviation) across the six random seeds are provided. Diff. denotes the relative improvement over the pinpoint baseline. Where applicable, the learned values of the $\sigma$ parameters are shown.}
    \label{tab:ablation}
    
    \begin{tabular}{
        p{0.13\textwidth}
        p{0.27\textwidth}
        p{0.15\textwidth}
        p{0.11\textwidth}
        p{0.22\textwidth}
    }
        \toprule
        \textbf{Method}
        & \textbf{Description}
        & \textbf{Accuracy (\%)}
        & \textbf{Diff. (\%)}
        & \textbf{Learned $\sigma$}
        \\
        \specialrule{\heavyrulewidth}{\aboverulesep}{\belowrulesep}

        Baseline
        & No redistribution; each predicted TRT value remains assigned to its original token
        & $69.23 \pm 4.92$
        & --
        & N/A
        \\

        \specialrule{\heavyrulewidth}{\aboverulesep}{\belowrulesep}

        Uniform 1:4
        & TRT is uniformly redistributed over one token to the left and four tokens to the right
        & $70.18 \pm 1.15$
        & $1.37$
        & N/A
        \\
        \midrule

        Fixed 1:4
        & Predetermined asymmetric Gaussian with a fixed 1:4 left-to-right spread
        & $71.46 \pm 2.28$
        & $3.22$
        & N/A
        \\
        \midrule

        Symmetric
        & A single learnable $\sigma$ parameter shared by the left and right sides
        & $74.19 \pm 1.86$
        & $7.16$
        & $\sigma = 3.28 \pm 0.84$
        \\

        \specialrule{\heavyrulewidth}{\aboverulesep}{\belowrulesep}

        Asymmetric
        & Independently learned left- and right-side $\sigma$ parameters
        & $74.81 \pm 1.32$
        & $8.06$
        & $\sigma_{\mathrm{left}} = 0.80 \pm 0.56$

          $\sigma_{\mathrm{right}} = 3.47 \pm 2.13$
        \\
        \bottomrule
    \end{tabular}
\end{table}

Lastly, the asymmetric Gaussian condition achieved the highest mean accuracy at 74.81\%. However, its advantage over the symmetric condition (74.19\%) was small relative to the variability across seeds. Hence, the strong performance improvement of the symmetric Gaussian indicates that the overall gain cannot be attributed solely to the asymmetric design of the redistribution.

\section{Discussion}
The current work has three main implications. First, the accuracy improvements in Table~\ref{tab:oasst1_results} suggest that gaze features are more informative when they are modeled as spatially distributed attention rather than as strictly token-local signals. Across all evaluation settings, the proposed redistribution approach outperforms the corresponding pinpoint gaze baseline, suggesting that treating gaze location as the center of an attentional distribution during text processing is a promising approach.

Second, the optimized asymmetric Gaussian parameters may serve as a bridge between reward modeling in LLMs and human attention modeling during reading. The fitted model shows rightward asymmetry across all experimental settings in our evaluation on OASST1 and UltraFeedback, with a larger average right-side width than left-side width (Table~\ref{tab:sigma_values}). This pattern is directionally consistent with the asymmetric perceptual span reported in left-to-right reading, where visual information is acquired from a broader region to the right of fixation than to the left. We further examine this correspondence through a comparison with the OB1-reader model; detailed procedures and results are provided in Appendix~\ref{app:ob1}. This suggests that the gaze-derived attentional distribution optimized for LLMs may reflect certain spatial properties of human attention, although this does not imply that the model provides a direct estimate of human attention. 

The results of the ablation study provide further insight into how this may operate. The symmetric Gaussian condition learns $\sigma=3.28$, close to the asymmetric Gaussian condition's right-side width ($\sigma_{\mathrm{right}}=3.47$), whereas the asymmetric condition learns a substantially smaller left-side width ($\sigma_{\mathrm{left}}=0.80$). This pattern suggests that information to the right of the fixated token contributes more strongly to preference prediction. A symmetric Gaussian with a width comparable to $\sigma_{\mathrm{right}}$ may capture a similar range of predictive information, but less efficiently because it necessarily distributes weight equally in both directions. 

Third, the proposed gaze redistribution method generally yields more stable performance across random seeds than the pinpoint gaze baseline, as reflected in the smaller standard deviations reported in Table~\ref{tab:oasst1_results}. We hypothesize that this stability improvement arises because pinpoint gaze features are sparse and token-local: a high TRT value assigned to a single token can disproportionately affect the learned representation, making the model sensitive to noise in the predicted gaze features and to seed-dependent optimization dynamics. The proposed redistribution mechanism smooths such information over neighboring tokens, reducing the influence of isolated noisy values and providing a denser representation.

Overall, these findings suggest that modeling gaze as a spatially distributed feature provides a useful extension to fixation-centric gaze features in reward modeling. More broadly, the current study highlights how grounding machine learning frameworks in established findings from cognitive science offers a principled way to improve model performance and stability.

\section{Limitations and Future Work}
We acknowledge several limitations of this work that provide avenues for future research. First, our redistribution mechanism remains a simplified approximation of human visual attention during reading. Although asymmetric Gaussian redistribution captures several important properties of the attentional distribution, including distance-based decay and left--right asymmetry, human attention is more complex and context-dependent. The attentional distribution during reading may vary depending on reading skill, text difficulty, typography, task demands, and linguistic context.

More importantly, the Gaussian kernel used in the current study imposes a monotonic decay from the fixation center. This assumption is useful as a simple approximation, but it may not fully capture the structure of visuospatial attention. For example, prior work has suggested that the attentional field may not simply decrease with distance, but can show a Mexican-hat-like pattern, where attention is enhanced near the attended location and suppressed in surrounding regions \citep{muller2005attentional}. Recent evidence also suggests that visual spatial attention can be nonmonotonic and may vary across individuals \citep{lee2022individual}. These findings imply that future gaze-augmented reward models may benefit from more flexible redistribution functions beyond a single asymmetric Gaussian, such as distributions with inhibitory surround effects, nonmonotonic profiles, or individually adaptive span parameters.

Additionally, the current implementation redistributes TRT values based on token-level distance. This is a design choice for compatibility with the ET prediction model, which predicts token-level TRT values. Future work could explore redistribution based on letter-level distance, which may better reflect the spatial scale of human visual attention.

Lastly, although TRT can serve as a proxy for the amount of attention allocated to each token, other gaze features not considered in the current study may provide complementary information. Future work may extend the framework to multiple gaze features, such as first fixation duration, fixation probability, and the number of fixations, which may capture different aspects of human visual attention.

\begin{ack} 
This work was supported by the New Faculty Settlement Research Program through Korea Advanced Institute of Science \& Technology (Grant No. G04240059 [to SHL]); the Data Science Convergence Human Resource Development Program through the National Research Foundation (NRF) of Korea funded by the Ministry of Science and ICT (Grant No. RS-2022-NR068758 [to SHL]); the AI Science Hub Program through the Ministry of Science and ICT (Grant No. N10260136 [to SHL]); and the Industry Research and Development Program funded by FMMENTALCARE (Grant No. G01250572 [to SHL]).
\end{ack}


\printbibliography

\newpage
\appendix
\begingroup
\section{Comparison with the OB1-reader Model}
\label{app:ob1}

To further examine whether the learned token-space redistribution aligns quantitatively with an established cognitive model of reading, we compare our learned asymmetric Gaussian kernel against simulated attentional trajectories from the OB1-reader model \citep{snell2018ob1, lopesrego2024language}. OB1-reader is a cognitive model designed to simulate how readers guide their gaze movements and allocate visual attention across words. Rather than processing text strictly one word at a time, the model assumes that readers process multiple neighboring words simultaneously, with the amount of visual attention allocated to each word decaying smoothly with distance from the attentional focus. Specifically, visual input across a five-word window is modulated by an asymmetric attentional gradient that extends further in the reading direction (i.e., to the right side).

\subsection{Experimental Setup}
For this analysis, we select the Llama-3-8B reward model under the GazeConcat setting on OASST1, as it achieved higher mean downstream accuracy than Llama-3-8B-Instruct across six random seeds (Table~\ref{tab:oasst1_results}). Across the six checkpoints, the learned parameter values averaged $\sigma_{\mathrm{left}} = 0.804$ and $\sigma_{\mathrm{right}} = 3.471$, corresponding to a 1:4.32 left-to-right width ratio. These parameters were optimized solely through downstream preference learning without any exposure to the Provo Corpus or OB1-reader simulations.

Using 100 OB1-reader trajectories generated on the Provo Corpus with the implementation provided by \citet{lopesrego2024language}, we project OB1-reader's letter-level attentional weights onto the relative T5-token positions used by our redistribution module.

In OB1-reader, the attentional weight $a_i$ assigned to a letter $i$ at a signed letter distance $f_i$ from the focus of attention is modulated by an asymmetric Gaussian distribution formally defined as follows \citep[Equation~2]{snell2018ob1}:

\begin{equation}
    a_i = \frac{1}{\mathrm{width}} \cdot \exp\left( - \frac{f_i^2}{2 \cdot (\mathrm{width} \cdot \mathrm{asym})^2} \right) + c_a.
    \label{eq:ob1_attention}
\end{equation}

We evaluate OB1-reader under the 1:4 left-to-right attentional asymmetry ($\mathrm{asym}=0.25$ toward the left and $\mathrm{asym}=1.0$ toward the right). This 1:4 configuration directly models established psychophysical estimates of the human perceptual span in left-to-right reading, where effective processing extends approximately four times further to the right of fixation (14--15 letter spaces) than to the left (3--4 letter spaces; \citealp{rayner2014gaze, snell2018ob1}).

\subsection{Comparison Conditions and Metrics}
We evaluate unit-TRT allocation across three conditions over OB1-reader's mapped five-word window (comprising one word to the left and three words to the right of the fixated word $n$ in the Provo simulation setup of \citealp{lopesrego2024language}):
\begin{itemize}
    \item \textbf{No redistribution (Pinpoint):} All weight remains strictly assigned to the directly fixated token without spatial spreading.
    \item \textbf{Symmetric Gaussian:} For each checkpoint, we construct a symmetric baseline with equal left and right Gaussian widths ($\sigma_{\text{left}} = \sigma_{\text{right}}$) calibrated to match the overall token-space spread of the learned asymmetric kernel. Because the five-word window (one word to the left and three words to the right of fixation) is itself asymmetrically bounded, equal Gaussian widths do not necessarily yield equal total weight on the two sides.
    \item \textbf{Learned asymmetric Gaussian (Ours):} Token-level redistribution using each checkpoint's independently learned parameter values $(\sigma_{\text{left}}, \sigma_{\text{right}})$ as defined in Equations~(1)--(2).
\end{itemize}

All comparison metrics are computed separately for each of the six random-seed checkpoints and then averaged. We report our results using two metrics: (1) Spearman rank correlation, which measures agreement in the relative ranking of token positions across the window, and (2) Jensen--Shannon (JS) divergence, which evaluates the discrepancy between the complete normalized attentional allocation distributions.

\begin{table}[h]
\centering
\caption{Quantitative comparison between redistributed token-level TRT under different conditions and attention simulated by OB1-reader on the Provo Corpus under the 1:4 left-to-right attentional asymmetry ($\text{asym}_{\text{left}} = 0.25$, $\text{asym}_{\text{right}} = 1.0$). Values represent the mean across six random seeds.}
\label{tab:ob1_comparison}
\begin{tabular}{lcc}
\toprule
\textbf{Redistribution Condition} & \textbf{Spearman Correlation ($\uparrow$)} & \textbf{JS Divergence ($\downarrow$)} \\
\midrule
No redistribution (Pinpoint) & 0.4645 & 0.2756 \\
Symmetric Gaussian & 0.8958 & 0.1584 \\
\textbf{Learned asymmetric Gaussian (Ours)} & \textbf{0.9095} & \textbf{0.1473} \\
\bottomrule
\end{tabular}
\end{table}

\subsection{Results and Analysis}
Table~\ref{tab:ob1_comparison} summarizes the results. Both Gaussian redistribution conditions correspond substantially better to OB1-reader's attentional allocation than the pinpoint baseline. Specifically, the symmetric Gaussian condition increases Spearman correlation from $0.4645$ to $0.8958$ and reduces JS divergence from $0.2756$ to $0.1584$. As for the learned asymmetric Gaussian, it outperforms the paired symmetric condition on both Spearman correlation ($0.9095$ vs. $0.8958$) and JS divergence ($0.1473$ vs. $0.1584$) under OB1-reader's $1:4$ asymmetry setting.

The numerical differences between the learned asymmetric and symmetric conditions are modest by design because their overall spatial spread was matched. However, they differ primarily in their directional left--right mass allocation. A breakdown of the pooled attentional weights across the left, center (fixated), and right regions of the window clarifies this effect:
\begin{itemize}
    \item \textbf{Center allocation:} The asymmetric and symmetric redistributions assign nearly identical weight to the fixated center position ($26.83\%$ vs.\ $26.69\%$, respectively).
    \item \textbf{Leftward allocation:} The asymmetric redistribution substantially reduces leftward weight from $26.65\%$ (symmetric) to $11.39\%$, bringing it much closer to OB1-reader's sharp leftward decay. 
    \item \textbf{Rightward allocation:} The asymmetric kernel assigns more weight toward the right side ($61.77\%$ vs. $46.66\%$ for the symmetric kernel).
\end{itemize}
These results suggest that Gaussian spatial spreading accounts for much of the shared local structure between redistributed TRT and attention simulated by OB1-reader, while the learned rightward asymmetry provides additional directional alignment with the $1:4$ attentional profile of OB1-reader.
\endgroup

\section{Experiment Details}
\label{app:experiment_details}

\subsection{Dataset Processing}
\label{dataset_processing}

We conduct experiments on the OpenAssistant Conversations dataset (OASST1; \citealp{kopf2023openassistant}), HelpSteer2 \citep{wang2024helpsteer}, and UltraFeedback \citep{cui2024ultrafeedback}. Following the reward modeling setup, each example is formatted as a prompt--chosen--rejected triplet, where the chosen response is expected to receive a higher reward than the rejected response. We use the English subset of the dataset, since the eye-tracking prediction model used in our experiments was trained on English reading data.

\subsection{Training Configuration}
\label{app:training_config}
\begin{table}[H]
\centering
\caption{Training configuration}
\label{tab:training_config}
\begin{tabular}{ll}
\toprule
\textbf{Hyperparameter} & \textbf{Value} \\
\midrule
Datasets & OASST1, HelpSteer2, UltraFeedback \\
Backbones & Llama-3-8B, Llama-3-8B-Instruct \\
Training epochs & 2 \\
Per-device batch size & 1 \\
Gradient accumulation steps & 8 \\
Optimizer & AdamW \\
Learning rate scheduler & cosine \\
Minimum learning-rate ratio & 0.7 \\
Weight decay & 0.1 \\
LoRA rank & 8 \\
LoRA alpha & 32 \\
LoRA dropout & 0.1 \\
Gaze projector dropout & 0.1 / 0.3 \\
Initial $\sigma_{\mathrm{left}}$ & 1.0 \\
Initial $\sigma_{\mathrm{right}}$ & 1.0 \\
Random seeds & 41, 42, 43, 44, 45, 46 \\
\bottomrule
\end{tabular}
\end{table}

\subsection{Evaluation Protocol}
\label{app:evaluation_protocol}

Reward modeling performance is evaluated using pairwise preference accuracy. For each prompt, the model assigns scalar reward scores to the chosen and rejected responses. A prediction is counted as correct when the chosen response receives a higher reward than the rejected response:

\[
r_\theta(x, y_w) > r_\theta(x, y_l).
\]
Accuracy is computed as the percentage of correctly ranked response pairs. This is the same evaluation criterion used in prior reward modeling work and in GazeReward, where accuracy is defined by whether the preferred response receives a higher reward score than the less preferred response.

\subsection{Computing Environment}
\label{app:computing_environment}
We trained the models on three local servers. 

The first server was equipped with an AMD Ryzen Threadripper PRO 5965WX 24-Core CPU, 503 GiB of RAM, and 4 NVIDIA RTX A6000 48 GB GPUs. 

The second server was equipped with an AMD Ryzen Threadripper PRO 7975WX 32-Core CPU, 251 GiB of RAM, and 1 NVIDIA RTX A5000 24 GB GPU. 

The third server was equipped with 2 Intel Xeon Gold 6226R 16-Core CPUs, 376 GiB of RAM, and 4 NVIDIA RTX A5000 24GB GPUs. 

Each training was done on a single GPU at a time.


\newpage

\end{document}